**Kenyan Sign Language (KSL) Dataset: Using Artificial Intelligence (AI) in Bridging Communication Barrier among the Deaf Learners**

**[1]Wanzare L, [2]Okutoyi J, [3]Kang'ahi M. & [3]Ayere M.**

[1]Department of computer science, [2]Department of special needs education & [3]Department of educational communication and technology
[1,2,3]Maseno University, Kenya

***Abstract***
*Kenyan Sign Language (KSL) is the primary language used by the deaf community in Kenya. It is the medium of instruction from Pre-primary 1 to university among deaf learners, facilitating their education and academic achievement. Kenyan Sign Language is used for social interaction, expression of needs, making requests and general communication among persons who are deaf in Kenya. However, there exists a language barrier between the deaf and the hearing people in Kenya. Thus, the innovation on AI4KSL is key in eliminating the communication barrier. Artificial intelligence for KSL is a two-year research project (2023-2024) that aims to create a digital open-access AI of spontaneous and elicited data from a representative sample of the Kenyan deaf community. The purpose of this study is to develop AI assistive technology dataset that translates English to KSL as a way of fostering inclusion and bridging language barriers among deaf learners in Kenya. Specific objectives are: Build KSL dataset for spoken English and video recorded Kenyan Sign Language and to build transcriptions of the KSL signs to a phonetic-level interface of the sign language. In this paper, the methodology for building the dataset is described. Data was collected from 48 teachers and tutors of the deaf learners and 400 learners who are Deaf. Participants engaged mainly in sign language elicitation tasks through reading and singing. Findings of the dataset consisted of about 14,000 English sentences with corresponding KSL Gloss derived from a pool of about 4000 words and about 20,000 signed KSL videos that are either signed words or sentences. The second level of data outcomes consisted of 10,000 split and segmented KSL videos. The third outcome of the dataset consists of 4,000 transcribed words into five articulatory parameters according to HamNoSys system. The dataset is significant to members of the deaf community in breaking communication barriers among deaf and promoting wider inclusion of the deaf in education and other sectors. The scalability of the innovation is key in empowering the deaf worldwide.*

## 1. Introduction

The deaf community is recognized as a linguistic minority group due to their unique mode of communication, which is utilized by a relatively small segment of the global population. According to the World Federation of the Deaf (WFD, 2013), there are approximately 72 million deaf individuals worldwide, with over 80% residing in developing nations. In Kenya, the 2019 census estimated the population at 50 million, including around 1 million deaf individuals, as reported by the Kenya National Survey for Persons with Disabilities conducted in 2008 (Jefwa, 2014). Jefwa regards the deaf as a linguistic minority as compared to those that use spoken language regardless of whether such languages are also minority languages in their own right. The numerical majority in the Kenyan population is formed by the hearing population thus leaving the deaf as a clear minority in terms of language and culture. Ninety eight percent of deaf children are born from families with hearing parents, creating a significant linguistic gap from birth (Jefwa, 2014). Most families lack proficiency in sign language, which results in deaf children being deprived of a language model within their immediate environment. This absence of accessible communication creates a substantial barrier between the deaf and hearing communities, impacting their social integration and development.

The challenges faced by deaf individuals in accessing society are significant. Hilde and Colin (2009) highlight that the absence of recognition for sign language, the lack of bilingual education, limited interpreting services and general lack of awareness about the deaf community restrict their societal participation. This segregation is often evident in educational, occupational and recreational settings, as noted by Butler, Skelton and Valentine (2001). Hochgesang (2015) discusses the perception of deaf individuals in Kenya, where they are frequently viewed as lacking intelligence and language skills, which further marginalizes them. Deaf children born to hearing families often face severe limitations in language access, leading to late school entry, sometimes between ages 7 and 19, which adversely impacts their academic progress. In environments where teachers do not use sign language, these children struggle to develop literacy skills beyond basic word recognition and simple sentences. This situation emphasizes the critical need for effective language acquisition strategies and educational adaptations to support deaf learners.

Hochgesang (2015) indicates that insufficient educational opportunities lead to low literacy rates, which often result in poor living conditions for deaf individuals. In her conclusion, Hochgesang (2015) notes that deaf people in Kenya face heightened risk of unemployment, poverty, and health issues.

In Kenya, the primary language of instruction in most schools is English, which has its own grammar that deaf learners might find difficult to comprehend in signed English. Deaf learners use Spatial sequential processing mode whereas hearing learners use Auditory sequential processing mode (Adoyo, 2004). According to the Gachathi (1976) report, Kenyan Constitution (2010) and Kenyan Sign Language Bill (2021), The deaf are supposed to be taught in Kenyan Sign Language which is their mother tongue and written English as medium of instruction. Because of the limited number of sign language interpreters, these learners cannot be placed in integrated classrooms in Kenya. Further, it is possible that the interpreters may misinterpret what the speaker is saying, thus losing information in the process of translating. To bridge this gap, the use of assistive Artificial Intelligence technology may be of importance. However, use of Artificial Intelligence technology in teaching learners with hearing impairment has been hardly explored in the African context.

The key question for this research is therefore, how can assistive AI technology be used in bridging language gap among deaf learners in Kenyan learning institutions? This highlights the necessity for developing a prototype assistive AI technology that can convert spoken English into Kenyan Sign Language in real-time, aiming to eliminate the language barrier between deaf and hearing learners during classroom interactions in integrated environments, thus fostering inclusion and lifelong learning. Consequently, this project aims to create an assistive AI application that translates both spoken and written English into real-time Kenyan Sign Language, helping to bridge the language gap for deaf learners in their everyday activities. It will take the input of spoken English sentences, convert them to text, then translate to KSL using language independent sign language notation and finally display or visualize the signs. The first objective will be to Build KSL dataset for spoken English and video recorded Kenyan Sign Language, then develop transcriptions of the KSL signs to a phonetic-level interface of the sign language. The output will be a KSL dataset that can be used to develop tools for Sign Language research. The methodology for developing the technology can further be recommended for adoption in Africa and worldwide in deaf education.

## 2. Background and Related work

### 2.1 Sign language

Deaf children do not have any innate problems with learning a language but have more difficulties learning a language that is spoken rather than visual. A study by Lorraine (2021) reveals that deaf babies born into signing rich families will acquire sign language easily and at the same pace of a hearing child born in a hearing family. Lorraine further notes that the fact that being deaf only impacts a child's ability to learn spoken language, it doesn't affect language acquisition. However, deaf children are most often surrounded by language role models (hearing people) they cannot emulate and learn spoken language from due to their inability to hear. This creates significant delays in early language development and intervention.

Kenyan Sign Language (KSL) is a visual language utilized by the deaf community in Kenya. Typically, when signing different signs and signals are used to form sentences. These signs can be broadly categorized as either manual or non-manual. Mweri (2018) notes that the major parameters or building blocks of Kenyan Sign Language just like in other sign languages include: Hand shapes or hand forms (articulators), movement or motion (manner of articulation), location (place of articulation), and palm orientation (manner of articulation). These four articulatory properties of KSL together are referred to as manual signs that are physically produced by the hands and other parts of the body. However, according to Mweri (2018), oftentimes manual signs have to combine with non-manual signs such as facial expressions, gestures and body language to make meaning. The above parameters can be viewed as the articulatory properties of a sign.

The deaf also use fingerspelling to articulate words that have no sign equivalent using letters of the alphabets (Baker, 2010). KSL has its own syntax, morphology, phonology, semantics and pragmatics Morgan (2022). It is a legitimate language like any other as it has its own vocabulary, community of speakers, duality and discreteness, its own grammar and its productive nature. Like any language, sign language also deals with morphology. In sign language morphology looks at how to combine meaningful sign components in a way to construct complex signs (Mweri, 2023).

In KSL, there are signs that are made up of single morphemes e.g. SUNDAY, CHILD etc. Such signs are morphologically simple and are called monomorphic signs. Signs that are made up of more morphemes (bound morphemes) are morphologically complex signs and are called polymorphemic signs. Mweri (2023) highlights the following examples of morphology for morphologically complex signs used in Kenyan sign language. One, the use of suffixes that can be used to derive nouns from verbs. An example is the word FARMER that is derived by combining the morphemes FARM + PERSON. The sign PERSON acts as a suffix (an agentive maker). The noun FARMER is derived from the verb FARM by attaching the suffix PERSON to it. Mweri (2023) also notes that most signs for professionals in KSL are derived from bounding morphemes since they add the suffix PERSON. The second is the use of prefixes. An example is in the signs for BELIEVE and AGREE which are signed as THINK^TRUE and THINK^YES respectively. The sign THINK acts as a prefix that is added to the root sign. Other examples are for words like BROTHER and HUSBAND that are made up of MAN^SAME and MAN^MARRY respectively.

## 2.2 Glossing

Glossing is the practice of writing a morpheme-by-morpheme 'translation' using English words (MacDonald, Donovan, & Everett 2023). KSL glossing involves writing each sign word or sentence in upper case to signify a word or a sentence that has been signed. The syntax of KSL sentences is different from the syntax of spoken languages like English. For example: English: *"the cat is under the table"* is glossed in KSL as: "TABLE CAT UNDER//". In the above example, the English sentence is unglossed and it follows Subject + Verb + Object word order. In KSL sentences, each sign word is glossed in upper case letters and it follows Object + Subject + Verb or sometimes Subject Object + Verb word order.

## 2.3 AI technologies for Sign language

Papasratis et al. (2021) conducted a survey on AI technologies designed to eliminate communication barriers for deaf or hearing-impaired individuals, significantly enhancing their social inclusion. The survey aimed to provide a comprehensive review of methods for capturing, recognizing, translating, and representing sign language, highlighting their benefits and drawbacks. Papastatis et al. (2021) note that datasets are crucial for the performance of methodologies regarding sign language recognition, translation, and synthesis and as a result a lot of attention is needed towards accurate capturing of signs and their meticulous annotation. Such datasets included: continuous sign recognition data-sets and isolated sign language recognition datasets. In this research, we will collect video recordings of KSL and the related spoken English sentences for both isolated alphabets and words and continuous sentences.

There has been an attempt to document videos on sign language. According to Bragg et al. (2019), videos can be used as a means of representing sign language that can make information on the signs easily accessible. However, pre-recorded videos face some challenges for instance the cost of producing quality videos is high, it is almost impossible to later modify the contents of the videos and the signers cannot remain anonymous (Kipp et al., 2011). That is why there is a shift towards using animated avatars as a way of presenting generated sign language (Elliott et al., 2008). Computer avatars can be adapted for different use cases and audiences. Also, animations can be adjusted dynamically if need be which allows for real-time use cases (Kipp et al., 2011).

Research has been done on AI systems that translate sign language to text or voice and the reverse, translating voice or text into sign language. For instance, Wen et al. (2021), in their study on sign language and sentence recognition, developed a system for translating sign language to text. They proposed an AI enabled sign language recognition and communication system that used wearable sensing gloves for detecting hand motions and a virtual reality interface for rendering the recognized signs, before translation the recognized sign into text or voice. The general glove solution was only able to recognize discrete single gestures (numbers, letters, or words) as opposed to complete sentences. The system would have allowed bidirectional and possibly remote communication between signers and non-signers. Similarly, Maria, Summaira, Ali and Haider (2021) carried out a literature review on sign language for Pakistan. They provide a framework for translating Pakistani sign language into speech and text while using convolutional neural networking (CNN). Wen et al. (2021) and Maria et al. (2021) both focused on conversion of sign language to text; however, our research focuses on the conversion of spoken language into sign language to enhance inclusion of the deaf learners.

Sign language production focuses on the translation of text or voice into a representation of the sign. For instance, Sannaulah et al. (2022) developed a technology called Sign4PSL for real-time automatic translation of text into Pakistan Sign Language (PSL), aimed at assisting deaf individuals. This system utilizes virtual signing characters to visually represent sentences. They emphasized that this lightweight, platform-independent application can function offline, addressing the communication barrier between hearing and deaf individuals. Building on their findings, the current project aims to create an assistive technology solution to help bridge the communication gap between the hearing community and the deaf population in Kenya. This project seeks develop an assistive Artificial Intelligence application that will translate spoken and written English text to real time Kenya Sign Language to bridge the language barrier among the deaf learners in their day-to-day activities. Little is known of any AI prototype that can translate speech

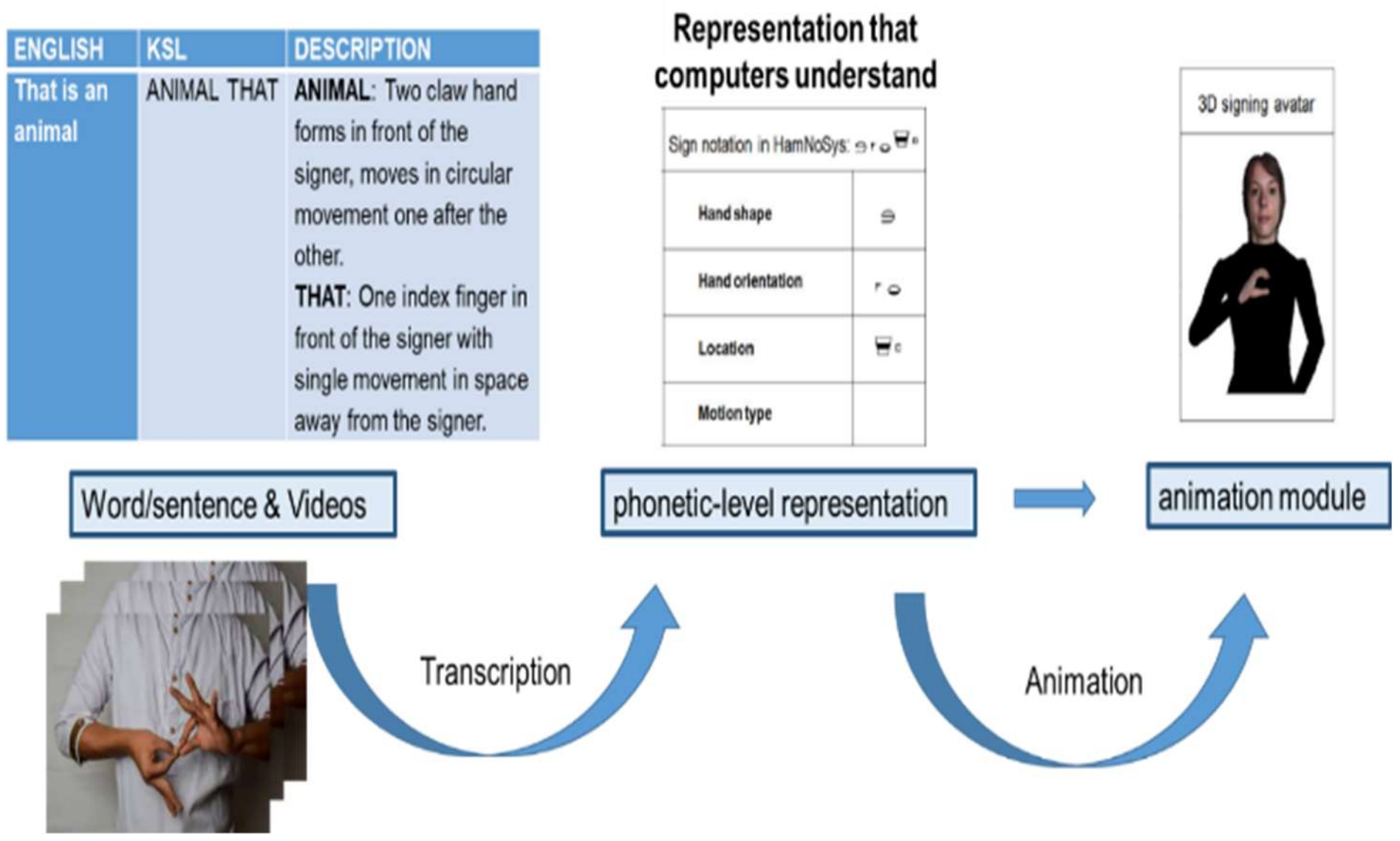

Figure 1: Figure summarizing the research methodology for the study (images adapted from Karpov et. al., 2016)

into Kenyan Sign Language. Thus, the need to develop a prototype for AI4KSL.

This project will provide an architectural framework of speech to Kenyan Sign Language. Secondly, AI4KSL technology will ensure full inclusion of deaf learners in the community by ensuring they access information, education and participate in day-to-day school and community activities.

When developing technology for translating spoken or text language to sign language, there are important modules that need to be considered. First speech recognizer as input if dealing with spoken language, the translation system that converts the text to sign language and finally a graphical avatar as output (Bungeroth et al., 2004; Khan et al., 2020). Our main concern is the translation system that converts text to sign language. Translation systems can be broadly categorized as either rule-based, corpus-based or neural machine translation (Kahlon et al., 2021). Rule-based systems for converting text to American sign language have been popular and more appropriate for building systems that convert text to sign language as there are normally no large datasets to train statistical or deep learning models (Huenerfauth, 2004; Kahlon et al., 2021). We follow a similar approach as to the best of our knowledge, there are no large datasets for KSL to training models.

## 2.4 Sign Writing Notation

The translation system requires sign writing notations that represent the various signs digitally. Various sign writing notations have been developed over time from Stokoe, Gloss and HamNoSys (Hutchinson, 2012). The HamNoSys (Prillwitz et al., 1989; Hanke, 2004, 2010) is a phonetic transcription system that has gained popularity due to its language independence, support for non-manual systems and lightweight nature (Sannaulah et al., 2022) and is also adopted in this research. HamNoSys uses ASCII and Unicode symbols thereby minimizing space when storing the signs in the computer (Hanke, 2004, 2010). From the translation output, the converted text should be rendered as an animation. HamNoSys itself does not have a representation for ease of rendering. HamNoSys can be converted to SiGML (Elliott et al., 2004) that is an XML framework that is an input to the animation module. There has been research on automatic ways of converting HamNoSys to SiGML (Kaur et al., 2016, Neves et al., 2020). Finally, the animation module renders the corresponding sign language represented by Signing Gesture Mark-up Language (SiGML) (Kipp et al, 2011; Yi et al., 2014). Yi et al. 2014 developed a prototype for an interfacing system as a communication medium platform for sign language users. The study proposed a sign language interfacing system that could be used to manage and edit sign language parts, by creating virtual human body parts and simulating virtual gestures.

## 2.5 Evaluation of sign language technology

The most commonly used models in evaluating assistive technologies are the Analyze, Design, Develop, Implement, and Evaluate (ADDIE) model by Kurt (2018), the CIPP model (Stufflebeam, 2003) and Kirkpatrick's Level model (Ruiz & Snoeck, 2018). Alodail (2014) used Kirkpatrick's model to evaluate the effectiveness of listening devices for technology rich learning environments for deaf learners. Ruiz & Snoeck (2014) also used Kirkpatrick's Kirkpatrick's model in evaluating assistive devices used in training and assessment of Technology Enhanced Learning. Research literature however maintains that CIPP and Kirkpractick's models are useful in evaluating assistive technologies for already existing innovations (Ruiz & Snoek 2018, Delee 2018 and Cercone & Poobsert 2013). O'keef (2012) and Kurt (2018) however explain that ADDIE is useful in evaluating universally designed assistive technologies for learning because it is useful in the instructional design process and also in the summative project evaluation.

This project as a result will adopt the ADDIE model because it will not only support the instructional design as used in the design thinking process but also as an evaluation strategy for the project (ELM Learning 2022, Kurt 2018, Naba 2022, & Vulpen 2015). As conversion of spoken language to sign language is considered a machine translation task, BLEU (Papineni et al., 2002), Word Error Rate (WER) (Ali et al., 2018), and Translation Error Rate (TER) (Snover et al., 2006) will also be used to intrinsically evaluate the system performance.

## 3. Research Methodology

### 3.1 Research Design

The project used mixed methods research within an experimental design (George 2021, Schoonenboom & Johnson 2017, and Plano & Ivankova, 2016) to capture both the qualitative and quantitative data needed in the design of an assistive Artificial Intelligence technology for Kenyan Sign Language (AI4KSL).

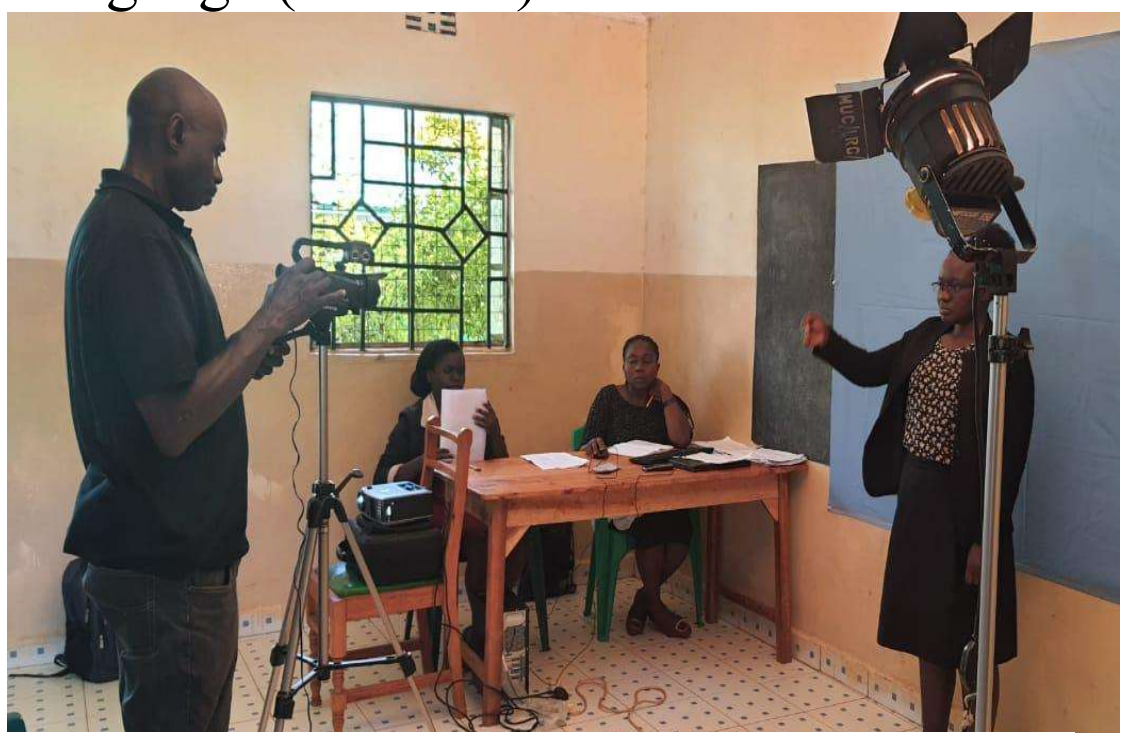

Figure 2: Video recording in session

### 3.2 Sample and sampling techniques

Stratified random sampling was used to select 21 sign language teacher trainees, 16 teachers of hearing-impaired learners and 600 learners who are deaf. The strata being the level of learners and class teachers for each cohort of learners from the early childhood level to class 8 primary school levels taking into account the gender of each respondent in cases. In addition, the sample was also obtained from one deaf boys' school, one girl's secondary deaf school, one mixed secondary deaf school and one tertiary mixed deaf technical training institution to ensure a heterogeneous and gender inclusive representative sample was used to get the various signs and sentences.

### 3.3 Dataset creation procedure

The following section describes the takes that were taken in the study to curate the sign language dataset. The dataset creation was broken down into two phases, the first phase involved the collection of the English sentences and corresponding KSL videos, while the second phase involved the transcription of the sentences and words into a phonetic-level representation using HamNoSys notation (Elliott et al., 2004; 2010).

The study borrows the methodologies outlined in the related work section. Figure 1 summarizes the methodology from collection of the text and video datasets, to the transcription process. The last phase is the conversion of the transcripts into a representation that can be rendered using an avatar. This last phase is left for future work.

### 3.4 Sentence collection

We curated a dataset of English sentences based on words collected from the KSL curriculum. The KSL curriculum is divided into several themes e.g. Home, Church, Court, Hospital etc., that are taught from primary to secondary level. We curated a total of about 6,000 sentences across the different topics.

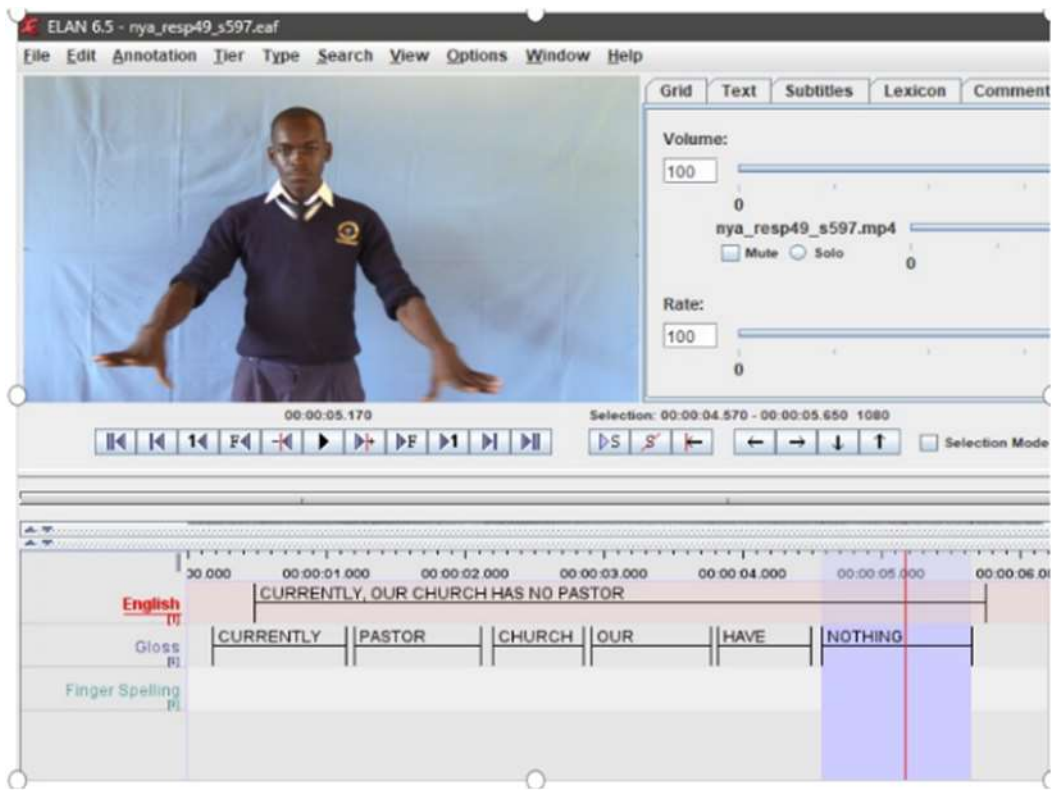

Figure 3: Example video editing using ELAN

Glossing

The collected KSL sentences were then glossed. Glossing is a common way of writing a morpheme-by-morpheme 'translation' of a given sign language using English words. KSL glossing takes the following syntax or word order: **Object + subject + verb or** sometimes **subject + object + verb.**

### 3.5 Sign video collection

This phase involved the development of a dataset of video recorded Kenyan Sign Language. Researchers collected videos of various signs in various subject domains offered in Kenya curriculum across various class levels in the deaf schools to meet KSL variations. The study incorporated at least 3 signers per sentence to capture sign variability. The study further ensured gender variations across schools. Data was collected through video recording using a front facing camera as shown in Figure 2.

### 3.6 Video preprocessing

After the video recording phase, a further cleaning and preprocessing was done. All videos were checked and split using Shortcut video editor to make sure that each video represented only a single word or sentence.

In a second preprocessing step, we used the ELAN tool to segment the videos, demarcating the start and end of each sign the video. The subsequent step involved defining Linguistic tiers within the ELAN tool, with three tiers established: English Sentence, Gloss, and Finger Spelling. Finger spelling needed to be captured in order to differentiate such sections from the other signs. Figure 3 shows an example video segmentation process using ELAN tool

### 3.7 Phase 2: Transcription

This phase involved the development of the transcriptions that would serve as input for developing the assistive technology. One main thing that is needed is a phonetic-level interface that describes sign language at the phonetic level. We used the HamNoSys notation (Hanke, 2004) which is a well-established language independent notation for the transcription of sign languages (Sanaullah et al., 2022; Dhanjal et al, 2020). Once the phonetic notation is developed, the study will convert the HamNoSys transcriptions into SiGML (Signing Gesture Markup Language) (Neves et al., 2020; Kaur et al., 2016; Elliott et al., 2004), which is an XML based annotation from which an animation can be rendered.

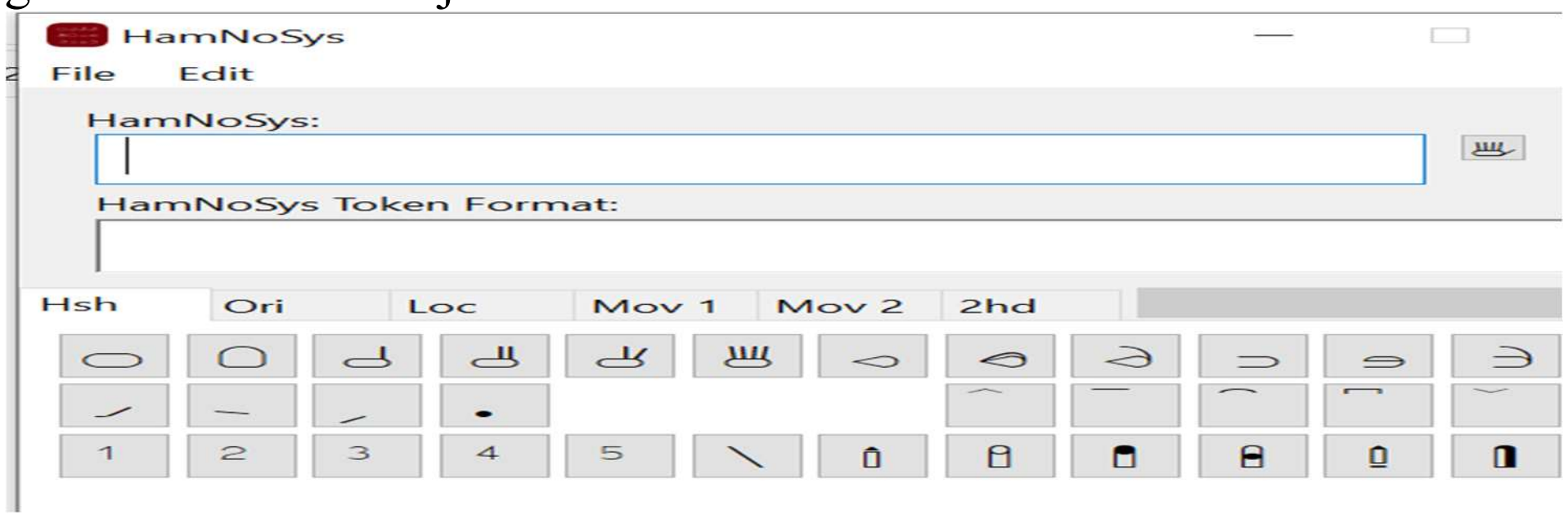

Figure 4: HamNoSys Pallete

Table 1: Example English sentences with their corresponding KSL Gloss

| | **English Sentence** | **KSL Gloss** |
|---|---|---|
| 1 | A bat always sleeps day time | ALWAYS DAY TIME BAT SLEEP// |
| 2 | A big lorry is blocking the way | NOW WAY LORRY BIG BLOCK// |
| 3 | A buffalo is a big animal | BUFFALO ANIMAL BIG// |
| 4 | A bullet can kill | BULLET KILL CAN// |
| 5 | A CAMEL HAS NICE MILK | MILK NICE CAMEL HAVE// |
| 6 | A CHAMELEON WALKS SLOWLY | CHAMELEON WALK SLOW// |
| 7 | A CHEETAH RUNS FAST | CHEETAH RUN FAST// |

## 4. Results and discussion

### 4.1 KSL dataset: Words and sentences

We curated a dataset of English sentences based on words from the KSL curriculum. We extracted about 4000 words from the KICD curriculum as a basis. From the words, research assistants involved in the study created about 6,000 sentences across different topics e.g. Home, Church, Court, Hospital etc.

In a second step, all sentences were translated into their corresponding KSL gloss. Example sentences and their corresponding glosses are shown in Table 1.

### 4.2 KSL dataset: Videos

The data were collected in two steps: first, a reconnaissance visit to each deaf school/college to meet the participants and acquaint them with the purpose of the research and distribute the KSL words and sentences data for practice before the actual data collection. Each respondent was required to sign 15 KSL words and 15 sentences. In the second step, the researchers visited each school and collected the KSL data through video-taping the KSL words and sentences signed by each participant. Each signed word and sentence were also audio-taped by the research assistant. Video sentences were segmented and imported into ELAN Software, saving them with reference to the assigned code name for each sentence.

### 4.3 KSL dataset: Transcription

All unique words that were collected were transcribed using HamNoSys notation. About 4,000 words were transcribed into the five articulatory parameters in HamNoSys. Transcription from English words to HamNoSys is done on Microsoft Excel spreadsheet.

The procedure for transcribing the words is presented below:

i. The HamNoSys Unicode software was installed which will allow the HamNoSys fonts to be recognized and displayed. The HamNoSysUnicode font was in Excel.

ii. To access the HamNoSys symbols for describing the signs, we used the offline HamNoSys palette. The HamNoSys Palette has six tabs containing all the symbols for describing signs put in different categories as shown in Figure 7.

iii. Each transcriber described the complete sequence of the HamNoSys symbols of

iv. a word in the HamNoSys text box in the HamNoSys palette, while maintaining the HamNoSys structure which begins with the handshape, hand orientation, palm orientation, location, and finally movement

v. The sequence of HamNoSys symbols, once fully described for a given word in the pallete, the symbols were copied and inserted in the HamNoSys Symbols column of the Excel spreadsheet for the respective word.

vi. The corresponding names of the HamNoSys symbols thus described were then copied from the HamNoSys Token Format text box of the palette and inserted on the HamNoSys Name column for the particular word. For every word, therefore, there is a column for the sequence of HamNoSys symbols and HamNoSys name.

## 5. Conclusion

In this paper, the study's aims, methodologies, and the related resources were presented. The paper describes the development of Kenyan Sign Language dataset consisting of 14,000 sentences with corresponding KSL gloss derived from a pool of about 4000 words from the KICD curriculum, and about 20,000 signed KSL videos. The second level of data outcomes consisted of splitting and segmenting the KSL videos, demarcating the beginning and ending of every sign together with a section of finger spelling. The third outcome of the dataset consists of 4,000 transcribed words into five articulatory parameters according to HamNoSys notation. The dataset can be used to foster further research into building assistive tools for people with hearing impairment.

| S/No. | English | GLOSS | HamNoSys Symbols | HamNoSys Name |
|---|---|---|---|---|
| 1 | ANCHOR | ANCHOR | [illegible] | hamsymmlr,hamfist,hamthumbacrossmod,hamextfingero,hamparbegin,hampalml,hamplus,hampalmr,hamparend,hamclose,hamchest,hammoved,hamtense |
| 2 | ANGEL | ANGEL | [illegible] | hamsymmlr,hamfinger2345,hamthumboutmod,hamparbegin,hamextfingerl,hampalmd,hamplus,hamextfingerr,hampalmd,hamparend,hamclose,hamshouldertop,hamwavy,hamrepeatfromstartseveral |
| 3 | ANCIENT | ANCIENT | [illegible] | hamfist,hamthumbacrossmod,hamextfingero,hampalml,hamchin,hamtouch,hammoved,hamwavy |
| 4 | ATMOSPHERE | ATMOSPHERE | [illegible] | hamfinger2345,hamthumboutmod,hamextfingero,hampalmd,hamheadtop,hamcirclei,hamrepeatfromstartseveral |
| 5 | ATTENTION | ATTENTION | [illegible] | hamsymmlr,hamflathand,hamparbegin,hamextfingero,hampalml,hamhead,hamlrbeside,hammoveo,hamplus,hamextfingero,hampalmr,hamlrbeside,hamhead,hammoveo,hamparend |
| 6 | PERSON | PERSON | [illegible] | hamflathand,hamparbegin,hamextfingero,hampalml,hamchest,hamlrbeside,hamplus,hamextfingero,hampalmr,hamlrbeside,hamchest,hamparend,hammoved,hamhalt |
| 7 | WRITE | WRITE | [illegible] | hamsymmpar,hamparbegin,hampinch12,hamdoublebent,hamextfingerol,hampalmd,hamfingertip,hamthumb,hamindexfinger,hamplus,hamflathand,hamextfingeror,hampalmu,hampalm,hamparend,hamclose,hamseqbegin,hammover,hamwavy,hamplus,hamnomotion,hamseqend |
| 8 | LOUD | LOUD | [illegible] | hampinchall,hamfingerstraightmod,hamextfingerl,hampalmd,hamlrbeside,hamneck,hammovel,hamtense,hamreplace,hamfinger2345,hamthumboutmod |
| 9 | LOUNDRY | CLOTHE AND WASH | [illegible] | hampinch12,hamfingerstraightmod,hamextfingeri,hampalmd,hamchest,hamlrat,hamcomma |
| 10 | LOW | LOW | [illegible] | hamfinger2,hamthumboutmod,hamextfingero,hampalmd,hamchest,hamlrbeside,hammoved |

Figure 5: Example English words, gloss and corresponding HamNoSys Symbol and name

## 6. Acknowledgements

This research was supported by EduAI Hub at the University of Lagos as part of a project under AI4D Africa. AI4D is a collaborative initiative by the International Development Research Centre (IDRC), Canada, and the Swedish International Development Cooperation Agency (SIDA), Sweden.

We also acknowledge the research assistants who took part in developing the dataset at various level as indicated below:

*Sentence Generation and Glossing* - Elsie Ochieng, Elizabeth Makhoha.

*Video Recording and Splitting* - Felix Okeyo, Harun Gad, the deaf Schools involved in data collection: Maseno School for the Deaf Primary, ACK Ematundu boys Secondary and Vocational School for the Deaf, Sikri Technical and Vocational training institute for Blind and Deaf, St Angela Girls Mumias Secondary and Vocational School for the Deaf, Fr Ouderaa Secondary School For the Deaf Nyangoma and Ebukuya Primary School For The Deaf Primary.

*Video splitting and segmentation* - Felix Okeyo, Harun Gad, Peter Kiprotich, Beverlyn, Elizabeth Makhoha, Samuel, Gilbert, Miriam, Kwamboka.

*HamNoSys transcription* - Godfrey Angwech, Elsie Ochieng, Sharifa, Brian, Rose, Sylvia, Mercy, George, Jackline, Sylvia.

## List of References